\documentclass{article}

\usepackage{arxiv}

\usepackage[utf8]{inputenc} 
\usepackage[T1]{fontenc}    
\usepackage{hyperref}       
\usepackage{url}            
\usepackage{booktabs}       
\usepackage{amsfonts}       
\usepackage{nicefrac}       
\usepackage{microtype}      
\usepackage{lipsum}
\usepackage{amsbsy}

\usepackage{subcaption}
\usepackage{graphicx}
\newtheorem{remark}{Remark}
\newtheorem{definition}{Definition}

\title{Unsupervised Space-Time Clustering using Persistent Homology}

\author{
  Umar ~Islambekov\\
  Department of Mathematical Sciences\\
  University of Texas at Dallas\\
  Richardson, TX 75080\\
  \texttt{umar.islambekov@utdallas.edu.edu} \\
   \And
  Yulia ~Gel \\
  Department of Mathematical Sciences\\
  University of Texas at Dallas\\
  Richardson, TX 75080\\
  \texttt{ygl@utdallas.edu} \\
}

\begin{document}
\maketitle

\begin{abstract}
This paper presents a new clustering algorithm for space-time data based on the concepts of topological data analysis and in particular, persistent homology. Employing persistent homology - a flexible mathematical tool from algebraic topology used to extract topological information from data - in unsupervised learning is an uncommon and a novel approach. A notable aspect of this methodology consists in analyzing data at multiple resolutions which allows to distinguish true features from noise based on the extent of their persistence. We evaluate the performance of our algorithm on synthetic data and compare it to other well-known clustering algorithms such as K-means, hierarchical clustering and DBSCAN. We illustrate its application in the context of a case study of water quality in the Chesapeake Bay.
\end{abstract}


\section{Introduction}\label{Intro}
	Clustering is one the most extensively studied problems in the unsupervised learning where the main goal is to partition a given dataset into subgroups or clusters so that the within-cluster points have more common features among themselves compared to those in other clusters. In contrast to the supervised framework, the unsupervised setting lacks the response variable associated with the features of observations to guide the learning process which makes the task more challenging.
	
	Abundance and diversity of space-time data naturally leads to myriad of learning methods in statistics, computer science and domain knowledge disciplines. Such diversity has resulted in the development of many clustering methods that are usually tailored to the nature of data at hand as well as to the disciplines they come from. Comprehensive reviews on clustering discussing various classes of algorithms, their advantages and scopes of applicability can be found in \cite{Clustering_survey}, \cite{Clustering_review}.
	
	The clustering problem for space-time processes often arises particularly in the context of studying ecological and environmental data. For example, in \cite{Distribution-Based_Clustering} a distribution-based clustering algorithm is used to discern taxonomic groups within complex microbial communities. In \cite{Bay}, the clustering algorithm TRUST with optimally chosen parameters is employed to identify the subgroups of water quality monitoring stations in the Chesapeake Bay. We return to this case study for more discussion in Section \ref{Application}.
	
	In many clustering problems, there is an inherit shape structure underlying the clusters which plays a key role in their formation. In this paper, we propose a new unsupervised clustering algorithm which utilizes the shape information associated with the clusters. To study and compare shapes we use \emph{persistent homology} as a tool from \emph{Topological Data Analysis} (TDA). TDA, rooted in algebraic topology and computational geometry, offers a flexible set of mathematical tools to study the topological and geometric aspects of data. The works of \cite{Edelsbrunner} and \cite{CZ} have laid foundation and gave forward momentum to the development of TDA-based methods \cite{TDAintro} with applications to diverse areas such as genomics \cite{B_cancer}, sensor networks \cite{Sensors}, shape recognition \cite{S-recognition}, brain diseases \cite{AmanmeetPD}. Using persistent homology in clustering is a novel approach \cite{Persistent_clustering}. One important aspect of persistent homology is the stability of its outputs against noise \cite{Stability_1, Stability_2, Stability_3}.
	
	There are two noteworthy TDA-based clustering algorithms: Mapper \cite{S-recognition} and ToMATo (Topological Mode Analysis Tool) \cite{Persistent_clustering}. See Section \ref{Description} for their main features and comparison to our algorithm.
	
	The paper is organized as follows. Section \ref{Methodology} discusses the common TDA methodology and approach to data analysis. Our clustering algorithm is described in Section \ref{Description}. The results of numerical experiments on synthetic data and comparisons to other clustering algorithms are presented in Section \ref{Experiments}. In Section \ref{Application} we apply the algorithm to study the clustering arrangement of the water quality monitoring stations in the Chesapeake Bay. The concluding remarks and the directions for future work are presented in Section \ref{Conclusion}.
	
	\section{Methodology}\label{Methodology}
	In the TDA setting, a given point cloud $\mathbb{X}$ is assumed to lie in some metric space (such as the Euclidean space) with an associated distance measure $d$. A choice of $d$ could be crucial as much of what follows in the analysis in terms of the results and conclusions is tied to it \cite{TDAintro}.
	
	As it is generally hard to extract desired information directly from a discrete set of points, we build upon the point cloud $\mathbb{X}$ an abstract simplicial complex in order to approximate the topology of the underlying structure or shape and thereby study its the topological and geometric properties. Neighboring graphs that are typically built on top of data capture only pairwise interactions, whereas simplicial complexes encompass higher order relationships among the data points \cite{TDAintro}.
	
	\begin{definition}
		Let $X$ be a discrete set. An \emph{abstract simplicial complex} is a collection $\mathcal{C}$ of finite subsets of $X$ such that if $\sigma\in\mathcal{C}$ then $\tau\in\mathcal{C}$ for all $\tau\subseteq\sigma$. If $|\sigma|=k+1$ then $\sigma$ is called a \emph{$k$-simplex}.
	\end{definition}
	
	There are several different types of simplicial complexes such as \emph{Cech}, \emph{Alpha} and \emph{Vietoris-Rips} complexes to name a few. An appropriate choice of a simplicial complex depends on several factors such as problem complexity, nature of data and computational cost. While some simplicial complexes have desirable theoretical properties, at the same time, they may suffer from being computationally inefficient or vice versa. Vietoris-Rips (VR) complex is one of the widely used types of simplicial complexes in applications for its easy construction and fast computational implementation.
	
	\begin{definition}[Vietoris-Rips complex]
		A \textit {Vietoris-Rips complex} at threshold (or scale) $\epsilon>0$, denoted by $VR_\epsilon$, is the abstract simplicial complex whose $k$-simplices, $k=0,\ldots, K$, consist of points which are pairwise within distance of $\epsilon$. If $X\subseteq \mathbb{R}^d$, a 0-simplex is identified with a point, a 1-simplex with an edge, a 2-simplex is a triangle and a 3-simplex is a tetrahedron.
		
	\end{definition}
	After equipping the point cloud with the simplicial complex structure, the desired topological and geometric properties are studied through the lens of \emph{persistent homology}. That is, we first build a family of nested simplicial complexes (called a \emph{filtration}) corresponding to an increasing sequence of thresholds. Second, we compute relevant topological summaries such as the number of connected components, tunnels (one dimensional holes) and cavities (two dimensional holes). Third, we track the behavior of these summaries or features as the threshold increases. The standard output of the above pipeline is the \emph{persistent barcode} which is a collection of stacked horizontal intervals (or bars) whose ends correspond to the appearance (or birth) and disappearance (or death) of respective features as the threshold increases. An equivalent summary to the persistent barcode is the \emph{persistent diagram} - a set of paired threshold values corresponding to the birth and death of each feature. The persistent barcode and diagram can be easily visualized. More importantly, they provide deeper insight into data. Namely, a feature that persists over a significant range of threshold values is likely to be a true feature, whereas the one which has a short life-span could be a by-product of noise. The idea of considering a whole range of threshold values, instead of attempting to find the optimal one, is the hallmark of persistent homology which has become a popular tool in TDA. Obviously, the extent of persistence for a feature to be classified as true one or not is determined by the problem. To address this issue, in \cite{Bootstrap} a bootstrap-based method was introduced to construct a confidence band for the persistent diagram to distinguish features. Alternatively, other methods utilize the stability of persistent diagrams for the construction of confidence sets \cite{Confidence_set}.
	
\begin{figure}
		\centering
		\includegraphics[width=5in]{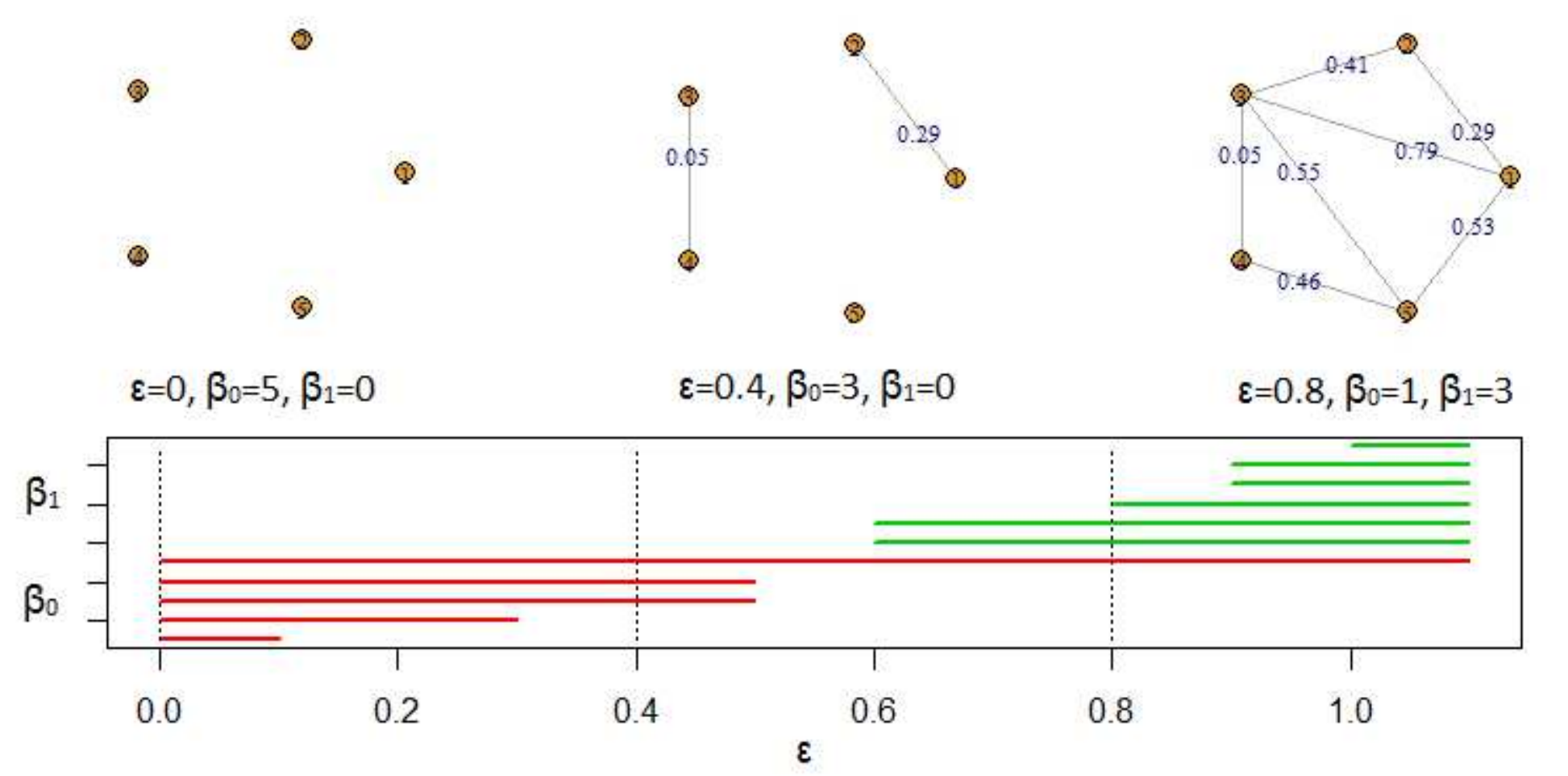}
		\caption{The persistent barcode of one dimensional Vietoris-Rips filtration built over five points. The top three figures are the snapshots of the evolving complex as threshold $\epsilon$ increases. Please note that the positions of the points in the figure do not exactly reflect the inter-point distances. The inter-point distances less than or equal to $\epsilon$ are depicted in blue. The $\epsilon$ values corresponding to the two ends of horizontal bars mark the birth and death of topological features. To find the Betti numbers we count the number of times respective horizontal bars intersect the vertical line through $\epsilon$. For example, for $\epsilon=0.8$, $\beta_0=1$ and $\beta_1=3$.}
		\label{fig:Barcodes}
	\end{figure}

	Another convenient way to encode the topological summaries provided by persistent homology is through the Betti numbers.
	\begin{definition}[Betti numbers]
		The $p$-th Betti number $\beta_p$, $p\in Z^{+}$, of a simplicial complex is the rank of the associated $p$-th \textit{homology} group defined as the quotient group of the \textit{cycle} and \textit{boundary} groups.
	\end{definition}
	In this paper, we do not delve into the mathematical details on the Betti numbers (see \cite{Hatcher} for more rigorous treatment). For our purposes, it suffices to mention that for a given complex, $\beta_0$ equals the number of connected components; $\beta_1$ the number of tunnels; $\beta_2$ the number of cavities, etc. The Betti numbers are closely tied to the persistent barcode and can be extracted from it (see Figure \ref{fig:Barcodes}).
	
		In applications, the extracted topological summaries are often forwarded to or integrated with other methods for further analysis of data \cite{TDAintro}. For example, the persistent diagram can be converted into the \emph{persistent landscape} which is more suitable for machine learning techniques to be directly applied \cite{Landscape}.
	
	There are standard libraries and packages for carrying out TDA related computations in C++, Python and R. Perseus is another noteworthy persistent homology software \cite{Perseus}. For more details on TDA, the reader may consult survey papers such as \cite{Carlsson}, \cite{TDAintro}, \cite{Ghrist}, \cite{WassermanTDA}.
	
	\section{Clustering using Betti Numbers (CBN) Algorithm}\label{Description}
	Let $\mathbb{X}$ be a point cloud consisting of $n$ points belonging to some metric space. Often $\mathbb{X}\subseteq \mathbb{R}^m$ and the components of the points represent spatial coordinates, quantitative features, or repeated measurements. Let $d$ be the associated distance function between the points.
	
	Our objective is to partition $\mathbb{X}$ into non-overlapping clusters, and our definition of a cluster falls under a general framework of density based clustering, e.g., DBSCAN, OPTICS etc (please see Section 4.5 of \cite{Clustering_survey}). That is, intuitively our cluster is to be a subset of points which is path-connected, i.e. any point in the subset can be reached from any other one through a path consisting of points (also belonging to the subset); furthermore, the consecutive points on the path are close enough and their local neighborhoods are similar in shape. To compare shapes, we build a Vietoris-Rips (VR) filtration upon such neighborhoods around each point and compute topological summaries in the form of the Betti numbers using persistent homology. The closer the Betti numbers to one another for a pair of close-by points, the more likely similar the shapes of their neighborhoods. Thus, when identifying clusters, CBN utilizes both the distance function and local geometric information around the points. Note that accounting for shape similarity can be viewed as an extension of conventional clustering properties in the density-based clustering framework. Below we summarize the main steps of CBN algorithm:
	\begin{enumerate}
		\item For each $i$ in $\mathbb{X}$, let $N(i)$ denote the neighborhood of point $i$ consisting of its $k$ nearest (with respect to the distance function $d$) neighbors (including itself). Let $D_i$ be the set of distances between all pairs of points in $N(i)$. Note that $|D_i|=k(k-1)/2$.
		\item Let $\hat{F}$ be the empirical cumulative distribution function (cdf) associated with the distances in all $D_i$'s. Transform these distances using a new distance function $d_{tr}=\hat{F}\circ d$ with range $[0,1]$ (see Remark \ref{remark2}).
		\item Fix an increasing sequence of non-negative thresholds $\epsilon_1<\epsilon_2<\ldots<\epsilon_l$. Let $VR_j(i)$ be VR complex corresponding to threshold $\epsilon_j$ whose $0$-simplices are all points in $N(i)$ and $1$-simplices are all pairs of points in $N(i)$ which are no more than $\epsilon_j$ distance apart with respect to $d_{tr}$. As a result, we get a filtration of VR complexes $VR_1(i)\subseteq VR_2(i) \subseteq \ldots \subseteq VR_l(i)$. By default, $\epsilon_j=0.01(j-1)$ and $l=100$.
		\item \label{step4} Compute the sequences of the Betti-0 and Betti-1 numbers, $\boldsymbol\beta_0^{(i)}=(\beta_{10}^{(i)},\beta_{20}^{(i)},\ldots,\beta_{l0}^{(i)})$ and $\boldsymbol\beta_1^{(i)}=(\beta_{11}^{(i)},\beta_{21}^{(i)},\ldots,\beta_{l1}^{(i)})$ on the corresponding VR filtration. Recall, $\beta_0$ counts the number of connected components of the complex, while $\beta_1$ represents the number of tunnels (or one dimensional holes).
		\item \label{step5} Update $N(i)$: keep only those $j$'s in $N(i)$ for which
		$$
			\frac{\|\boldsymbol\beta_0^{(j)}-\boldsymbol\beta_0^{(i)} \|}{\|\boldsymbol\beta_0^{(i)}\|}\leq \tau_0\ \ \mbox{and} \ \ \frac{\|\boldsymbol\beta_1^{(j)}-\boldsymbol\beta_1^{(i)} \|}{\|\boldsymbol\beta_1^{(i)}\|}\leq \tau_1,
		$$
		where $\| \cdot \|$ is the Euclidean norm and $\tau_0$ and $\tau_1$ are tunning parameters serving as upper bounds on relative change in the Betti numbers. Thus, the Betti numbers help refine $N(i)$, retaining only those nearest points whose local neighborhoods are relatively similar in shape to that of $i$. The extend of relative similarity is determined by $\tau_0$ and $\tau_1$.  		
			
		\item Form an adjacency matrix $A=\{a_{ij}\}_{n\times n}$, where $a_{ij}$ = 1 if and only if $j$ is in (the updated) $N(i)$. Compute the strongly connected components of the directed graph defined by $A$. The connected components are our clusters. In other words, points $i$ and $j$ are in the same cluster if and only if there is a path in each direction between them. Optionally, the clusters can be defined as weakly connected components of the undirected graph induced by $A$. In that case, the resulting clusters, in general, tend to be relatively larger in size.
	\end{enumerate}
	
	Our algorithm does not require knowing the total number of clusters to be identified, which is of particular importance in many applications involving clustering of environmental space-time processes. Optionally, if a lower bound on the number of clusters or the minimum number of points each cluster should contain is known, CBN algorithm can reassign the points that fell into the unwanted small clusters using the concept of depth function, e.g. Mahalanobis depth \cite{Mahalanobis}: assign a point to the cluster with respect to which its depth is the largest.
	
	\begin{remark}\label{remark1}
		In our studies, the VR complexes are typically built on small dense local point clouds (of size 5-12). In this case, if the dimension of the complexes is bounded by two, there would be almost no one-dimensional holes and many two-dimensional ones. However, by limiting their dimension to one (i.e., the VR complexes contain only zero and one-dimensional simplices), we allow the one-dimensional holes to be formed by three or more points (i.e., the triangles are not "filled" because they are not regarded as two-dimensional simplices) and therefore their number will be abundant. Nevertheless, the algorithm is flexible enough to include simplices of dimension two and higher if needed.		
	\end{remark}
	
	\begin{remark}\label{remark2}
		The persistent homology software Perseus \cite{Perseus}, used to compute the Betti numbers for our studies, requires that the thresholds increase with constant increment. This may be considered undesirable in light of the fact that the distribution of original distances is often skewed. However, the transformed distances (using $\hat{F}$) are approximately uniformly distributed on $[0,1]$, which paves the way to choose the thresholds to be evenly spaced. Observe that choosing the thresholds in this way is equivalent to selecting them non-uniformly on the scale of original distances accounting for regions where they are dense and sparse. Moreover, note that since $\hat{F}$ is a non-decreasing function, $N(i)$ contains the same $k$ nearest neighbors regardless of which distance function, $d$ or $d_{tr}$, is used.
	\end{remark}

	\begin{remark}\label{remark3}
		There is no a priori best way to choose the number of nearest neighbors $k$ and the tunning parameters $\tau_0$ and $\tau_1$ which directly impact the performance quality of the algorithm. Therefore, when choosing their values one has to rely on subjective judgment using, for example, the boxplots or quantiles of all the relative changes in the Betti numbers. To help guide the selection of $\tau_0$ and $\tau_1$, by default they are initially set to the maximum of relative changes in $\beta_0$ and $\beta_1$ respectively after outlying values (those which are larger than the upper quartile plus 1.5 times the inter-quartile range) are dropped or equivalently the upper whiskers in the corresponding boxplots (see Section \ref{Experiments} for an illustration). Though, this method for choosing default values of the tuning parameters $\tau_0$ and $\tau_1$ yields satisfactory clustering results in our numerical studies presented in this paper, it is ultimately up to the user to decide what their appropriate values should be. 	
\end{remark}

\begin{remark}
	The Betti numbers are closely tied to the persistent barcode. In fact, one can extract the Betti numbers from the corresponding persistent barcode. The persistent diagram is an alternative representation of the information contained in the barcode. We know that if two point clouds have similar topological/geometric properties, their corresponding persistent diagrams of Cech or VR filtrations are also close (with respect to the Bottleneck distance). Thus, if the Bottleneck distance between the persistent diagrams is large, then the point clouds are topologically/geometrically distinct. We conjecture that a similar result should be true for the Betti numbers. However, we have not come across any theoretical result addressing this issue in the existing literature.
\end{remark}
		
		In Section \ref{Intro} we briefly mentioned two TDA-based clustering algorithms: Mapper and ToMATo. Both of them rely on a \emph{filter function}, $f:\mathbb{X}\rightarrow \mathbb{R}^d$, to build simplicial complexes. A density estimator, centrality and eccentricity functions are among the common choices of filter functions in applications \cite{TDAintro}. Both Mapper and ToMATo act in conjunction with another clustering algorithm. In this regard, Mapper is very flexible and any clustering algorithm can be used with it. In ToMATo, $f$ is taken as an approximation of the density function from which data points are sampled. Then its persistent diagram is calculated and used to enhance the graph-based hill-climbing algorithm of \cite{Koontz}. In contrast, CBN does not depend on a filter function nor another clustering algorithm thereby differing in its approach to clustering.	
		
		Tackling the sensitivity of our algorithm's performance to the globally defined tuning parameters is still an open problem. The default way for selecting $\tau_0$ and $\tau_1$ by the algorithm (see \ref{remark3}) is rather heuristic and needs a theoretical justification. Additionally, if there are many "bridges" between clusters, then CBN may have difficulty separating them. We plan to address these issues in our future research. Moreover, for large datasets, computing the Betti numbers associated with each point becomes computationally very intensive. In that case, one can run CBN on a subset of the dataset and assign the remaining points to the resulting clusters using an appropriate distance/proximity measure.

	\section{Numerical experiments}\label{Experiments}
\begin{figure}
		\centering
		\includegraphics[width=3in]{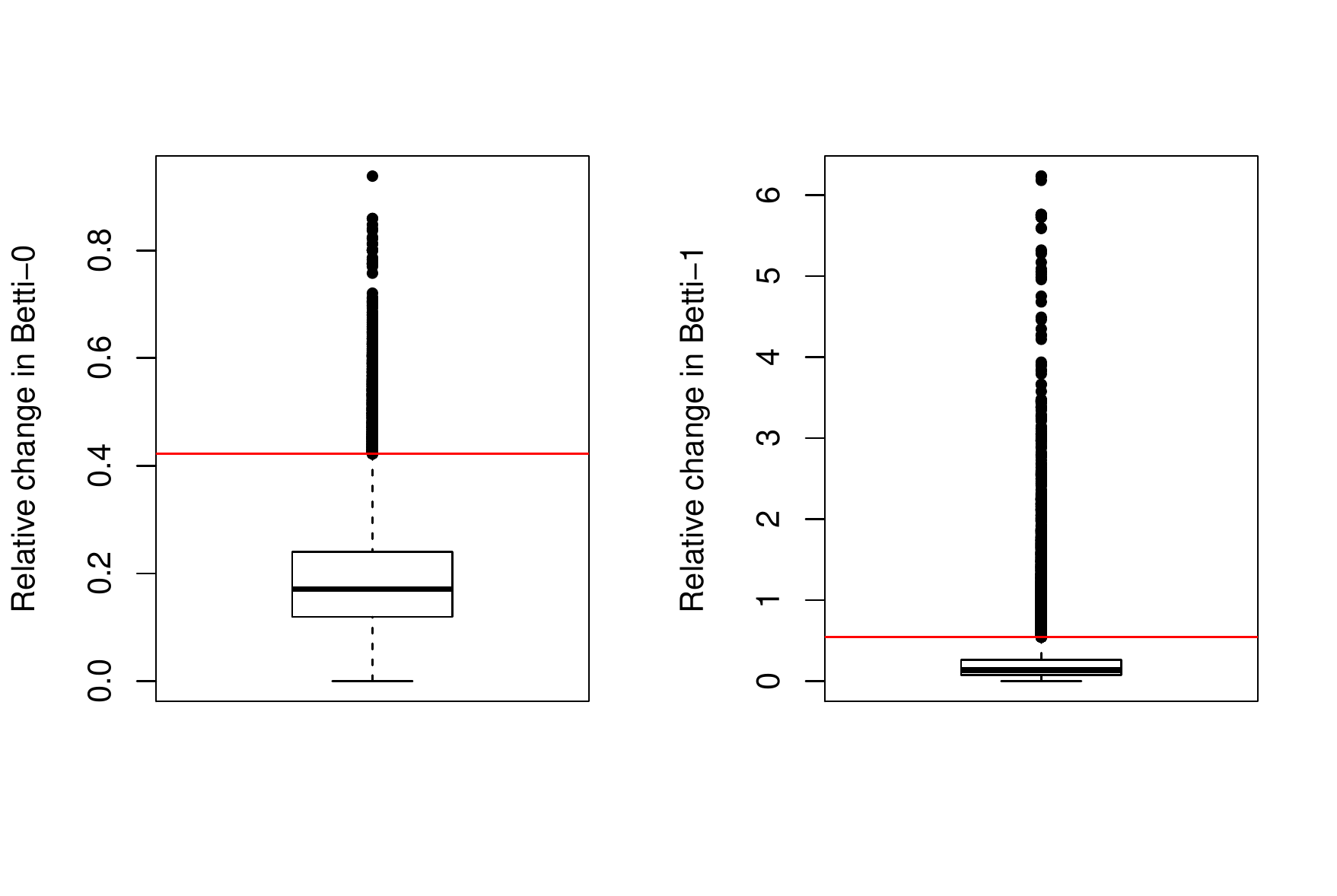}
		\caption{Relative changes in $\beta_0$ and $\beta_1$ numbers. By default, $\tau_0$ and $\tau_1$ are set to equal the upper whiskers (0.42 and 0.54 respectively). The outlying values are depicted as black points.}
		\label{whiskers}
	\end{figure}

	We now test our algorithm on a synthetic dataset. The dataset consists of 3800 points in $\mathbb{R}^2$ and 13 clusters with distinct shapes. The distribution of the points within the clusters is mostly uniform. We take $k=12$ (i.e., the size of neighborhoods around each point) and $\tau_0=0.42$ and $\tau_1=0.54$ (i.e., the tunning parameters as selected by the algorithm, see Figure \ref{whiskers} ).
	
\begin{figure}
		\centering
		\includegraphics[width=3in]{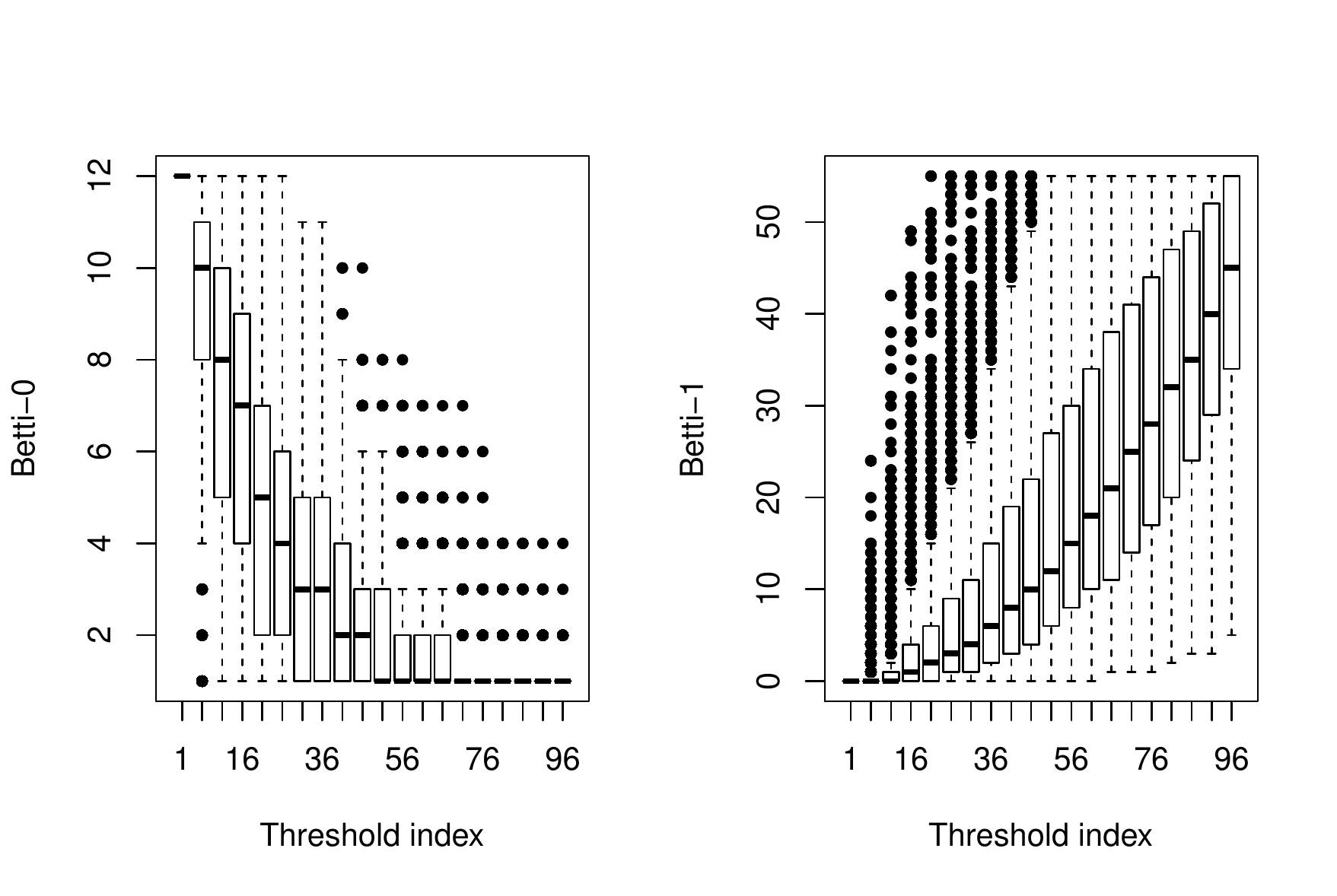}
		\caption{Box-plots of $\beta_0$ and $\beta_1$ numbers for increasing threshold values.}
		\label{boxplots}
	\end{figure}

	As threshold values increase, the corresponding VR complexes contain more and more 1-simplices (or edges) which results in a quick decrease of the number of connected components ($\beta_0$) and a gradual increase of one-dimensional holes ($\beta_1$) (see also Remark \ref{remark1}). This dynamics of $\beta_0$ and $\beta_1$ numbers is illustrated in Figure \ref{boxplots}.
	
\begin{figure}
		\begin{subfigure}{.5\textwidth}
			\centering		
			\includegraphics[width=2in]{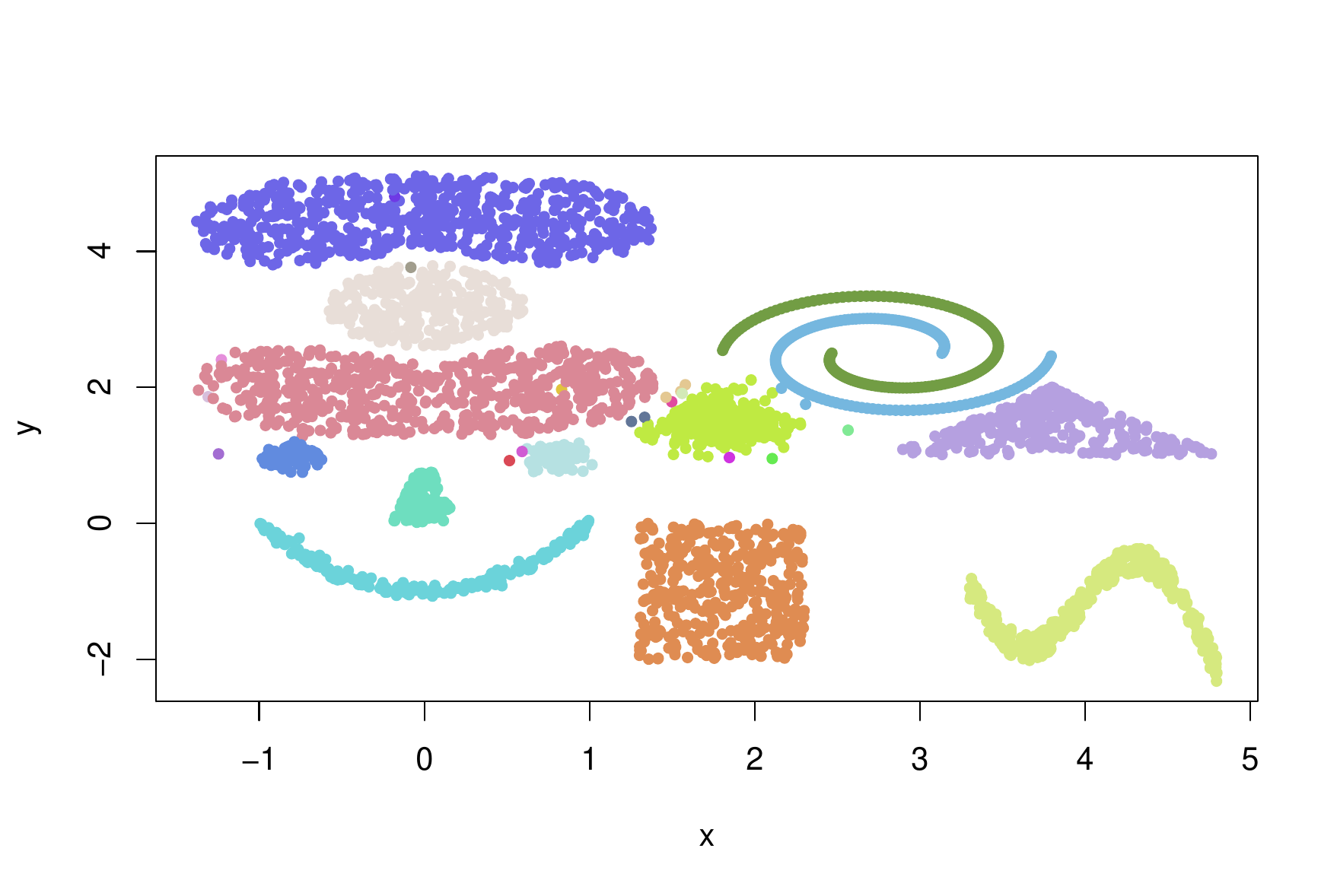}	
			\caption{CBN}
			\label{CBN}
		\end{subfigure}
		\begin{subfigure}{.5\textwidth}
			\centering
			\includegraphics[width=2in]{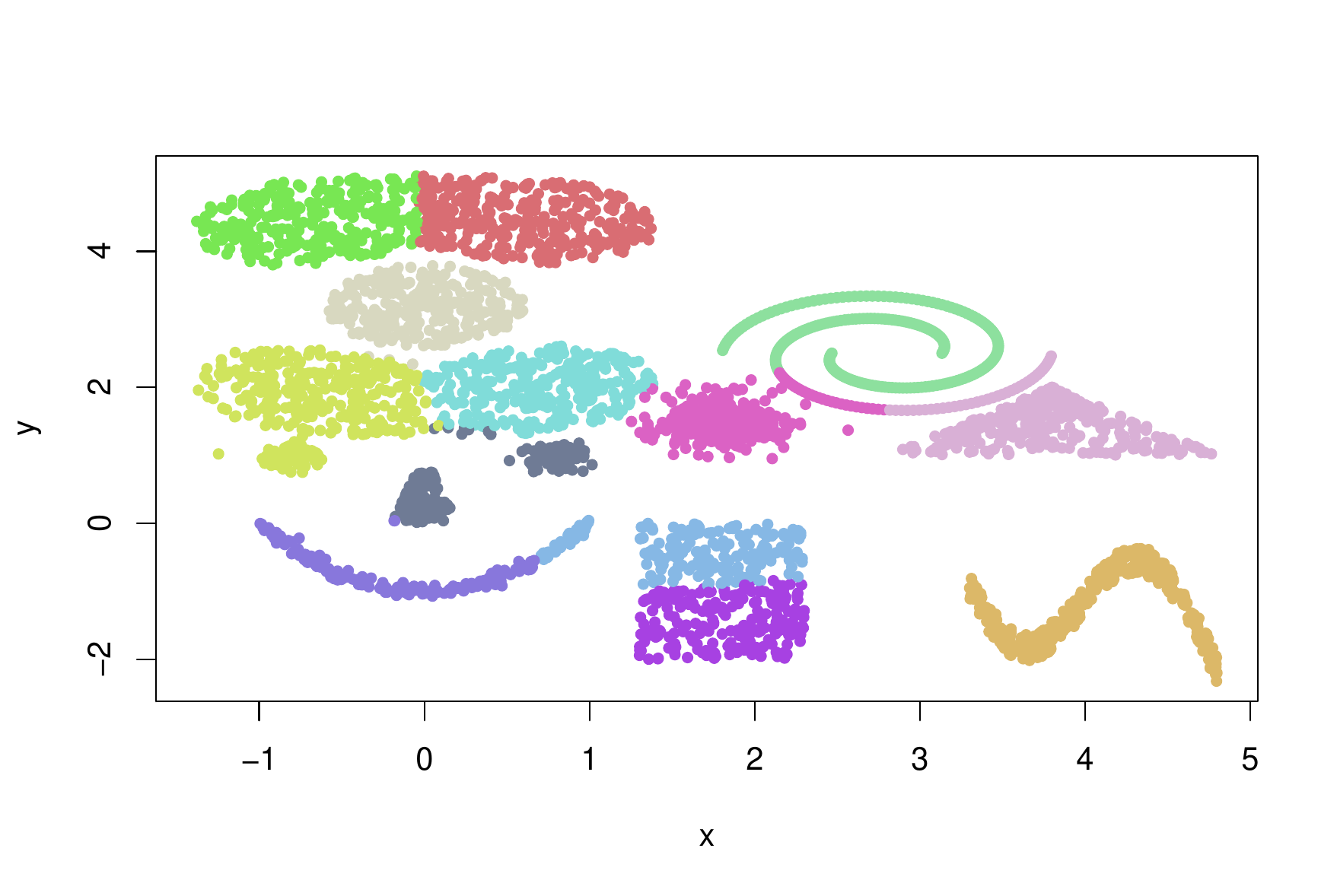}	
			\caption{K-means}
			\label{K-means}
		\end{subfigure}
		\begin{subfigure}{.5\textwidth}
			\centering
			\includegraphics[width=2in]{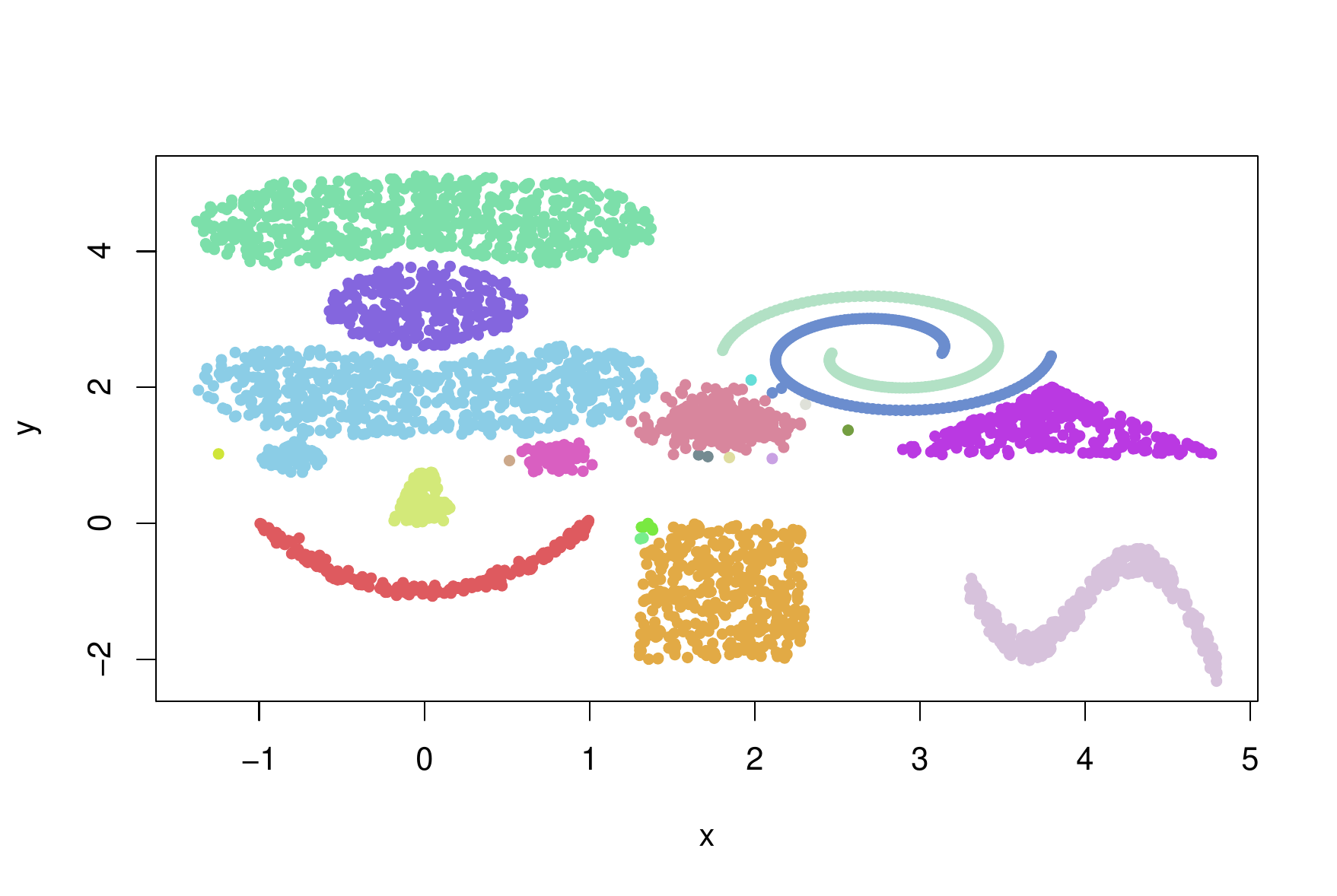}	
			\caption{Hierarchical}
			\label{Hierarchical}
		\end{subfigure}	
		\begin{subfigure}{.5\textwidth}
			\centering
			\includegraphics[width=2in]{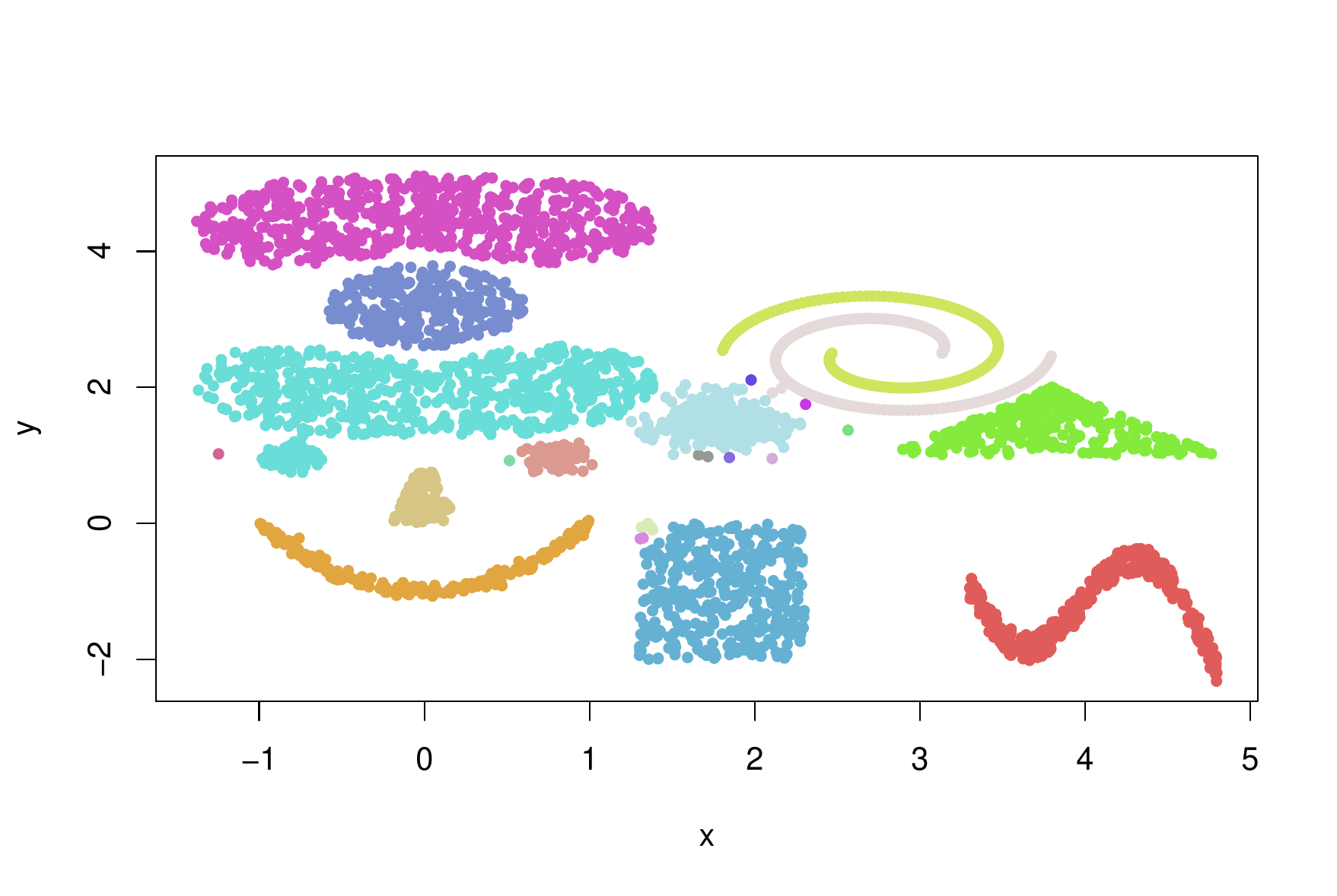}	
			\caption{DBSCAN}
			\label{DBSCAN}
		\end{subfigure}	
		\caption{Performace of CBN (Clustering using Betti numbers), K-means, Hierarchical and DBSCAN.}
	\end{figure}

	The performance of CBN algorithm is depicted in Figure \ref{CBN}. As Figure \ref{CBN} indicates, the algorithm delivers very accurate performance. Observe that despite the closeness of some clusters to each other, CBN is able to distinguish them successfully.
	
	Now, we compare the performance of CBN with K-means, hierarchy-based algorithms and DBSCAN which belong to different classes of clustering algorithms. K-means partitions data into clusters so that the within-cluster variance is minimized \cite{K-means}. The algorithms based on hierarchy build a hierarchical relationships among data points to perform clustering \cite{Hierarchy}. DBSCAN is a density-based algorithm which forms clusters around regions with high density \cite{DBSCAN}. Note that K-means requires pre-specifying the number of clusters, whereas hierarchical clustering algorithms and DBSCAN do not. For hierarchical clustering we used the \emph{single, complete} and \emph{average linkage} methods. The best performance is attained by the \emph{single linkage} method. The optimal cutoff parameter is selected based on the dendrogram obtained from hierarchical clustering. Their performance is illustrated in Figures \ref{K-means}-\ref{DBSCAN} and evaluated using Rand and Jaccard indexes \cite{Rand}, \cite{Jaccard}, denoted by $RI$ and $J$ respectively (see Table \ref{RJ}), and defined as:
		$$
		RI=\frac{TP+TN}{TP+FP+FN+TN}\ \ \mbox{and} \ \ \ J=\frac{TP}{TP+FP+FN},
		$$	
	where $TP$ is the number of true positives; $TN$ is the number of true negatives; $FP$ is the number of false positives; $FN$ is the number of false negatives. Table \ref{RJ} indicates that for the given dataset, among the four algorithms, CNB outperforms the rest. It is important to note that if steps \ref{step4} and \ref{step5} of CBN algorithm (see Section \ref{Description}) were skipped i.e., the neighborhood information was not updated using the Betti numbers, five major clusters would be lumped into a single cluster (not shown here). In contrast, K-means shows the poorest performance. Single linkage hierarchical clustering algorithm and DBSCAN perform sufficiently well except for failing to distinguish the 'right eye' cluster from the one above it.
	
	\begin{table}[h]
		\centering
		\caption{Summary of clustering performance for CBN (Clustering using Betti numbers), K-means, Hierarchical and DBSCAN based on Rand and Jaccard indexes.}
		\begin{tabular}{ lc c }
			\hline
			& Rand index & Jaccard index\\
			\hline\hline
			CBN & 0.99885 & 0.98880 \\
			K-means & 0.95661 & 0.62042 \\
			Hierarchical & 0.99225 & 0.92945 \\
			DBSCAN & 0.99225 & 0.92945 \\ 	
			\hline
		\end{tabular}
		
		\label{RJ}
	\end{table}

	Next, we add 200 uniformly distributed random points to the dataset in order to evaluate robustness of the four algorithms in the presence of noise.
	
	\begin{figure}
		\begin{subfigure}{.5\textwidth}
			\centering
			\includegraphics[width=2in]{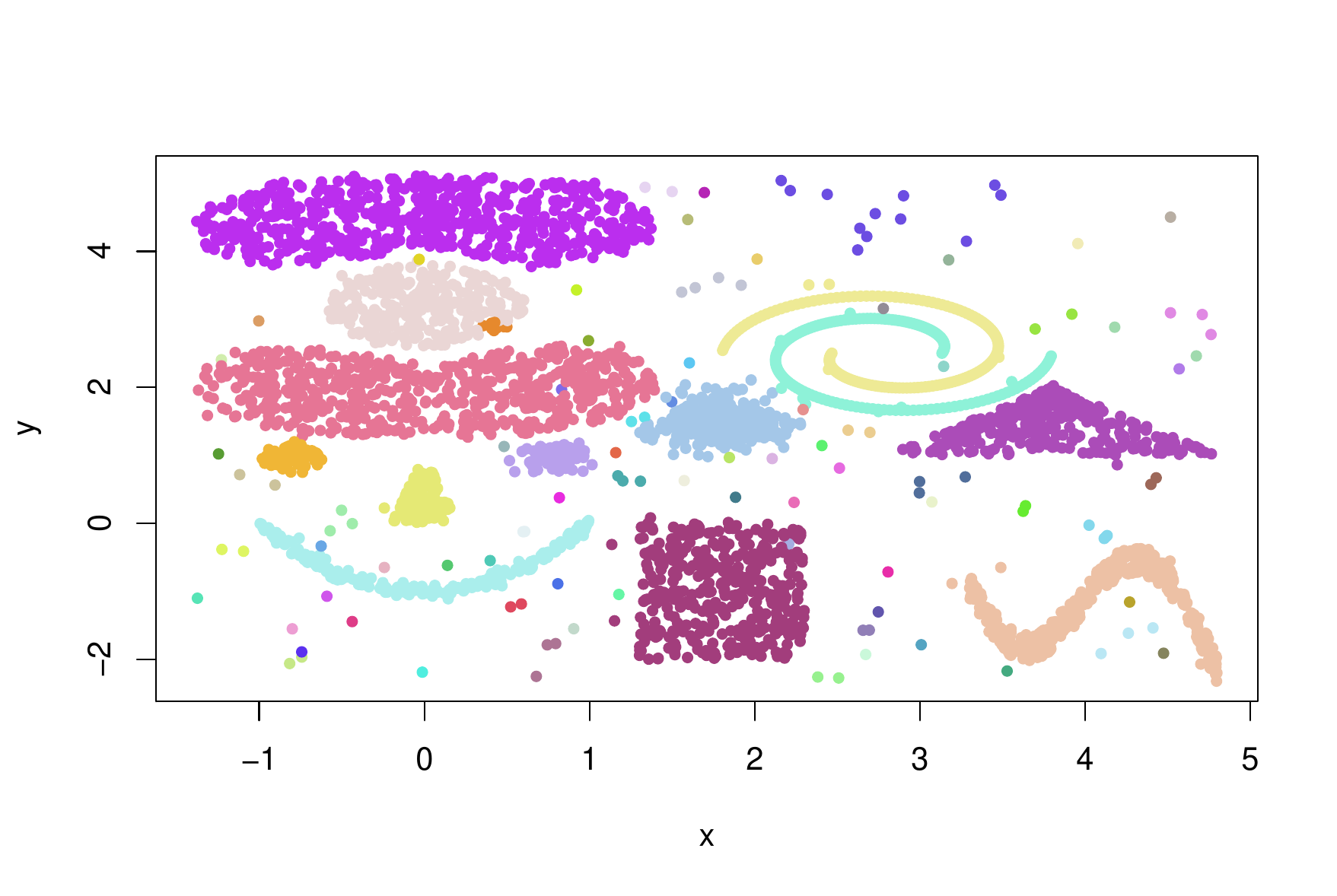}	
			\caption{CBN}
		\end{subfigure}
		\begin{subfigure}{.5\textwidth}
			\centering
			\includegraphics[width=2in]{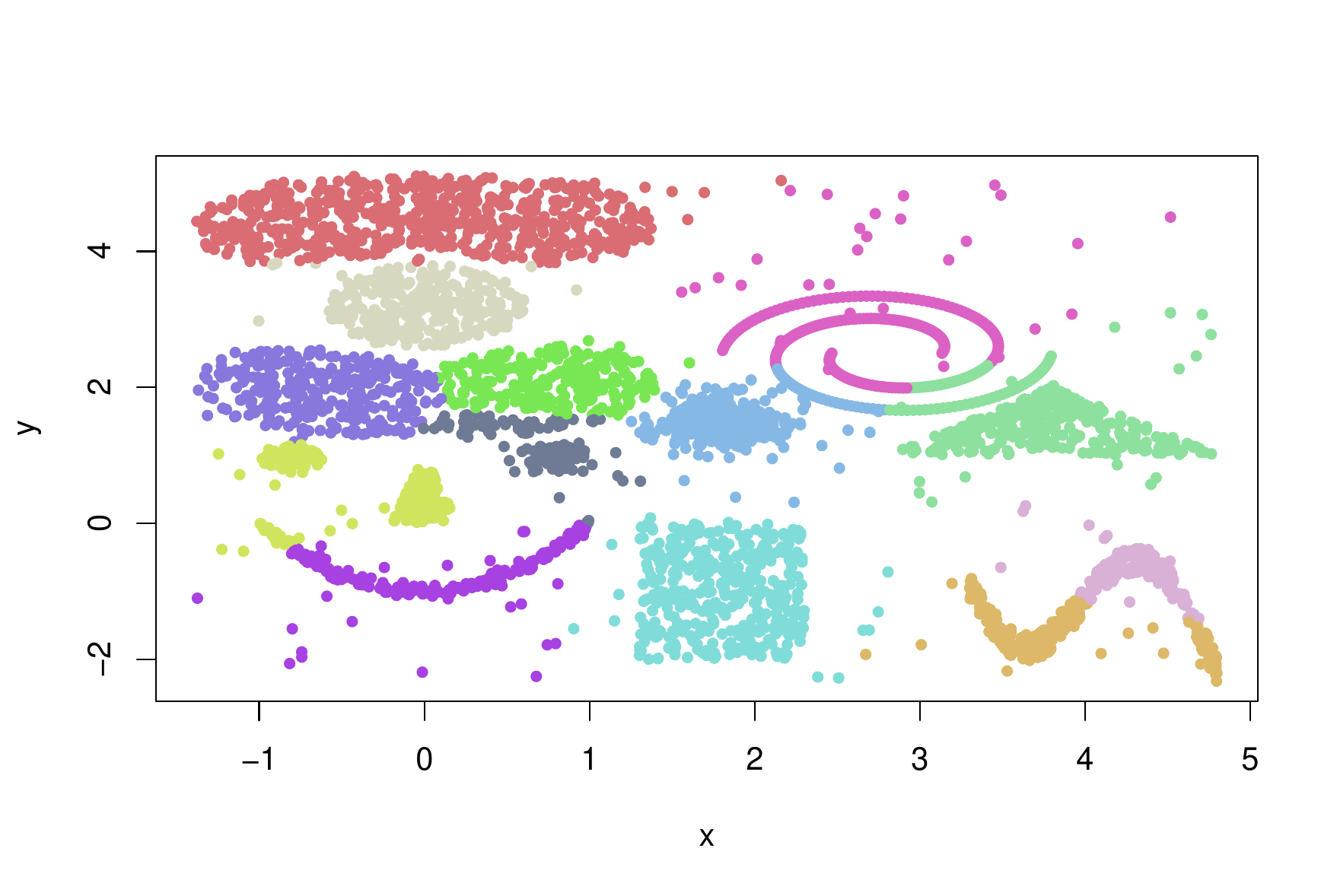}	
			\caption{K-means}
		\end{subfigure}
		\begin{subfigure}{.5\textwidth}
			\centering
			\includegraphics[width=2in]{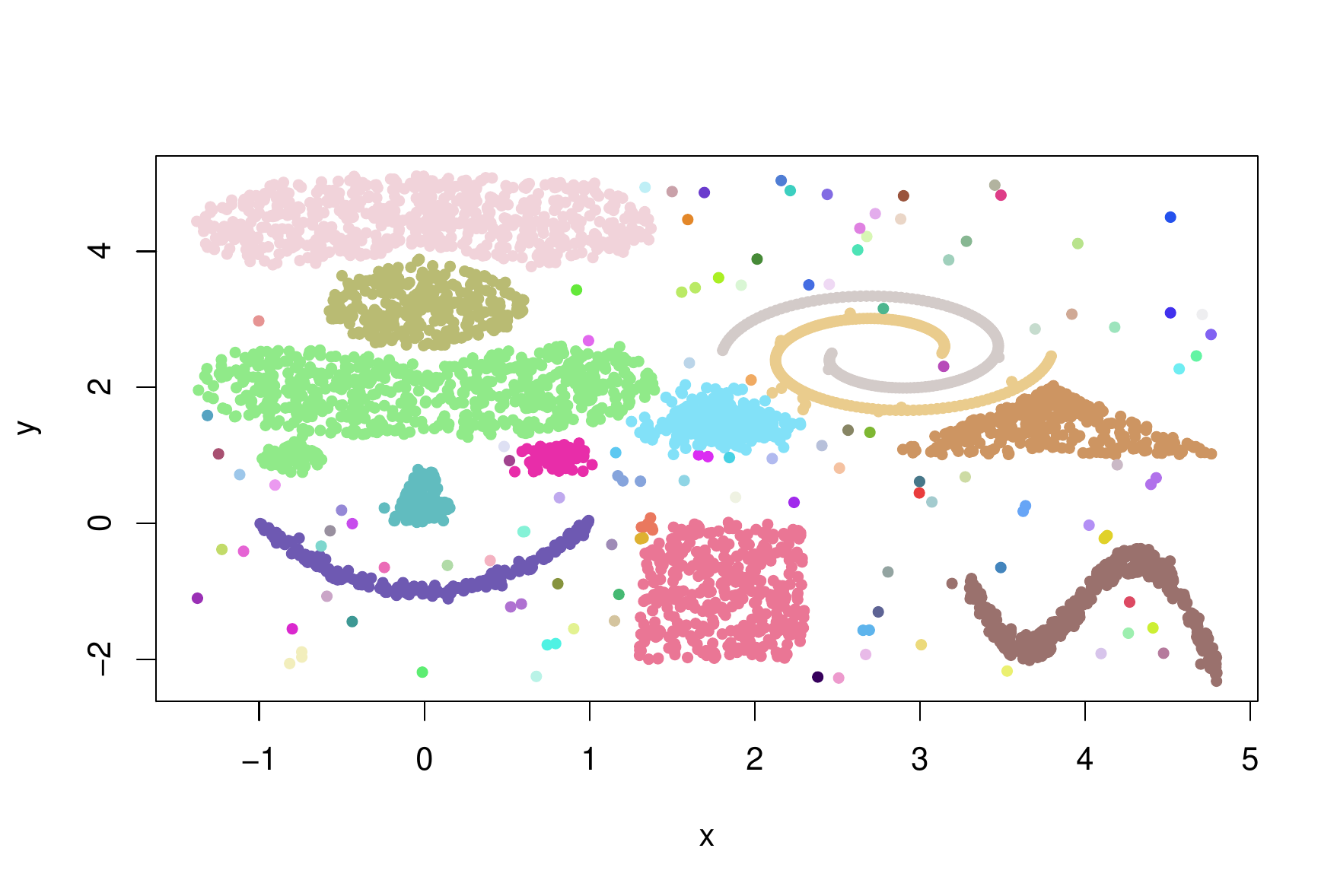}	
			\caption{Hierarchical}
		\end{subfigure}	
		\begin{subfigure}{.5\textwidth}
			\centering
			\includegraphics[width=2in]{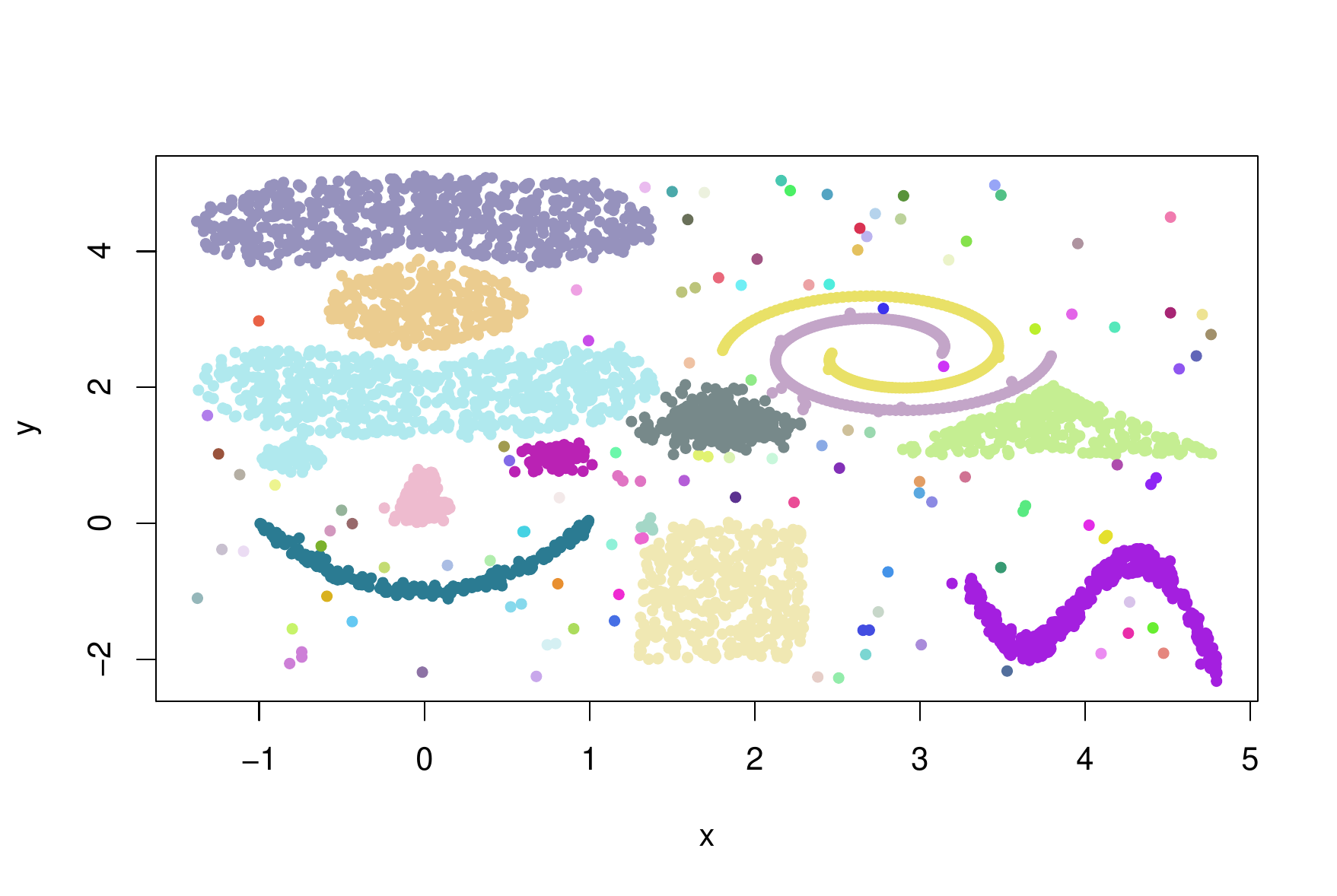}	
			\caption{DBSCAN}
		\end{subfigure}	
		\caption{Performance of CBN, K-means, Hierarchical and DBSCAN in the presence of noise.}
	\end{figure}
	Again, CBN algorithm correctly identifies all 13 clusters with few mismatches. K-means, Hierarchical clustering and DNSCAN show similar performance as before.

	\section{Case study - clustering of water stations in the Chesapeake Bay }\label{Application}
	In this section, we apply our method to analyze spatio-temporal data of water quality in the Chesapeake Bay during the period from 1985 to 2016. In light of facing threatening water pollution problems in the Chesapeake Bay, legislative regulations were passed in 1983 to monitor, protect and restore the bay ecosystem, in particular the water quality. As part of the measures, more than 300 stations were installed throughout the Bay area to collect water samples and monitor their pollution level over time.
	
	We analyze the publicly available data containing various water quality parameters recorded during 1985-2016 in the Chesapeake Bay. We consider 133 selected stations and use the level of suspended solids or sediment (TSS) as an indicator of water quality. The data are split into two 16-year periods - 1985-2000 and 2001-2016 to separate a possible effect of the Chesapeake 2000 agreement which introduced a new set of restoration measures. Measurements are combined into monthly averages; missing values are filled in using data from nearby stations.
	
	The same dataset is analyzed in \cite{Bay} using clustering algorithm TRUST with optimally chosen parameters. Following the dataset preprocessing of \cite{Bay} we scale the measurements for each station to have zero mean and unit variance to focus on trends (rather than particular TSS values).
	
	In the context of this problem, the set $\mathbb{X}$ consists of 133 stations. Each station has associated $12\times 16=192$ TSS measurements for the two periods. The distance measure $d$ is taken to be the Euclidean distance between the scaled TSS measurements. The size of neighborhoods around each station, $k$, is set to 7. We use the default values of the tunning parameters $\tau_0$ and $\tau_1$ for both periods.
	
\begin{figure}
		\centering
		\includegraphics[width=4in]{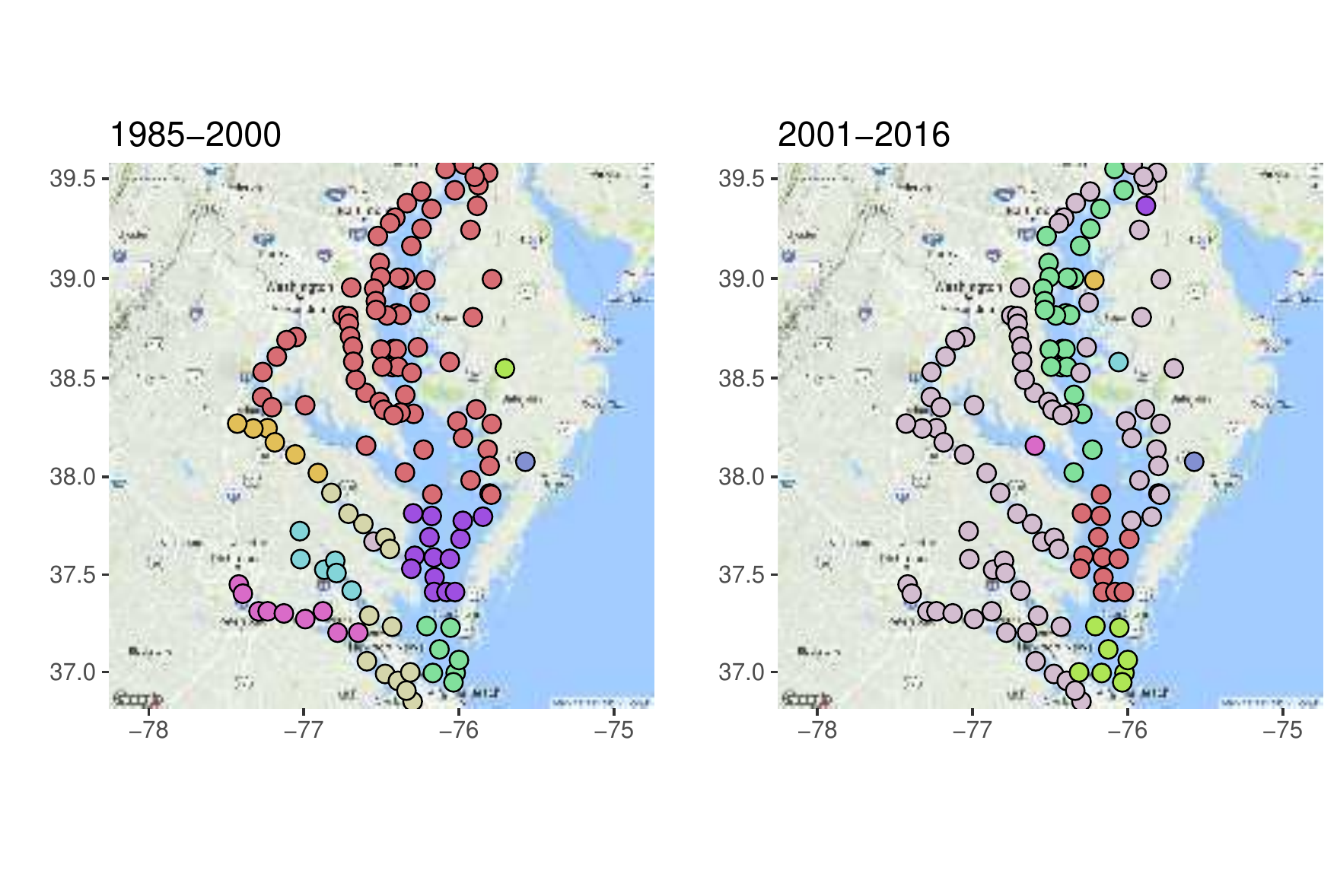}
		\caption{Clustering of water quality monitoring stations in the Chesapeake Bay using CBN algorithm.}
		\label{Chesapeake}
	\end{figure}

	Figure \ref{Chesapeake} shows the performance of CBN algorithm for the two periods. For 1985-2000 period, CBN identifies seven main clusters. The dominant cluster includes almost all the northern stations. The remaining clusters are formed around the stations which are spatially close to each other. This is noteworthy since no spatial information is supplied to the algorithm. In 2001-2016 period, we observe some shift in how clusters are formed. The largest cluster primarily consists of the stations located in the bay tributaries such as James, York, Rappahannock, Potomac, Gunpowder and Manokin rivers. It is known that the restoration efforts during 2001-2016 were concentrated only in the rivers. In view of this fact, the trend similarity among the stations in the largest cluster appears to be a result of these efforts. The other three main clusters are more towards the middle of the Bay. Note that two of these clusters are almost the same for both periods. The Rand index between the two clustering results is 0.63819.
	
\section{Conclusion}\label{Conclusion}
In this paper we have introduced a new space-time clustering method based on the concept of persistent homology. The key benefits of our approach are that we systematically account for data shape and geometry, and do not require a-priori knowledge on the number of clusters. In our numerical studies we have shown that our new method outperforms the competing approaches by providing more accurate clustering results for clusters of varying shapes and density. In the future we plan to extend our approach to unsupervised learning of complex networks and dynamic clustering of space-time data. Integrating the Betti numbers as extracted topological summaries into other clustering algorithms to boost their performance is another avenue worth exploring.

\section*{Acknowledgement}
The work is supported in part by the National Science Foundation under Grants No. IIS 1633331 and NSF ECCS 1824716.

\bibliographystyle{unsrt}  
\bibliography{references}

\end{document}